\documentclass[english,twoside]{article}
\usepackage[T1]{fontenc}
\usepackage[utf8]{inputenc}
\usepackage{array}
\usepackage{multirow}
\usepackage{amsmath}
\usepackage{amsthm}
\usepackage{amssymb}
\usepackage{graphicx}
\usepackage[numbers]{natbib}
\PassOptionsToPackage{normalem}{ulem}
\usepackage{ulem}

\makeatletter

\providecommand{\tabularnewline}{\\}

\usepackage[accepted]{aistats2018}

\usepackage{booktabs}
\usepackage{amsfonts}
\usepackage{microtype}
\usepackage{authblk}
\usepackage{textcomp}
\usepackage{clrscode3e}
\usepackage{enumitem}
\setlist{leftmargin=5mm}

\newcommand{\texttildemiddle}{\raisebox{0.5ex}{\texttildelow}}

\newcommand{\Requires}{\kw{assert}\,\Indentmore}
\newcommand{\Ensures}{\kw{assert}\,\Indentmore}

\renewcommand{\If}{\kw{if}\,\Indentmore}
\renewcommand{\Else}{\kw{else}\,\Indentmore}

\renewcommand{\For}{\kw{for}\,\Indentmore}

\newcommand{\Pa}{\mathrm{Pa}}
\newcommand{\Ch}{\mathrm{Ch}}
\newcommand{\Fo}{\mathrm{Fo}}

\usepackage{tikz}
\usetikzlibrary{positioning}

\tikzstyle{node}=[circle,draw,minimum size=0.7cm]
\tikzstyle{m-node}=[node,label={[label distance=-0.1cm]30:\tiny M}]
\tikzstyle{t-node}=[node,thick,label={[label distance=-0.1cm]30:\tiny M,T}]
\tikzstyle{i-node}=[node,thick,dotted,label={[label distance=-0.1cm]30:\tiny I}]
\tikzstyle{r-node}=[node,dashed,black!40,label={[label distance=-0.1cm]30:\tiny\color{black!40}R}]
\tikzstyle{edge}=[-]
\tikzstyle{stem}=[edge,ultra thick]

\definecolor{altblue}{cmyk}{0.8, 0.3, 0, 0.1}

\newcommand{\N}{\mathcal{N}}

\author[1]{Lawrence~M.~Murray}
\author[2]{Daniel~Lundén}
\author[1]{Jan~Kudlicka}
\author[2]{David~Broman}
\author[1]{Thomas~B.~Sch\"{o}n}
\affil[1]{Uppsala University}
\affil[2]{KTH Royal Institute of Technology}

\makeatother

\usepackage{babel}
\begin{document}
\twocolumn[\aistatstitle{Delayed Sampling and Automatic Rao–Blackwellization of Probabilistic Programs}

\aistatsauthor{Lawrence~M.~Murray \And Daniel~Lundén \And Jan~Kudlicka}
\aistatsaddress{Uppsala University \And  KTH Royal Institute of Technology \And Uppsala University}

\aistatsauthor{David~Broman \And Thomas~B.~Sch\"{o}n}
\aistatsaddress{KTH Royal Institute of Technology \And Uppsala University}
]

\runningauthor{Murray, Lundén, Kudlicka, Broman, Sch\"{o}n}
\begin{abstract}
We introduce a dynamic mechanism for the solution of analytically-tractable
substructure in probabilistic programs, using conjugate priors and
affine transformations to reduce variance in Monte Carlo estimators.
For inference with Sequential Monte Carlo, this automatically yields
improvements such as locally-optimal proposals and Rao–Blackwellization.
The mechanism maintains a directed graph alongside the running program
that evolves dynamically as operations are triggered upon it. Nodes
of the graph represent random variables, edges the analytically-tractable
relationships between them. Random variables remain in the graph for
as long as possible, to be sampled only when they are used by the
program in a way that cannot be resolved analytically. In the meantime,
they are conditioned on as many observations as possible. We demonstrate
the mechanism with a few pedagogical examples, as well as a linear-nonlinear
state-space model with simulated data, and an epidemiological model
with real data of a dengue outbreak in Micronesia. In all cases one
or more variables are automatically marginalized out to significantly
reduce variance in estimates of the marginal likelihood, in the final
case facilitating a random-weight or pseudo-marginal-type importance
sampler for parameter estimation. We have implemented the approach
in Anglican and a new probabilistic programming language called Birch.
\end{abstract}

\section{INTRODUCTION}

Probabilistic programs extend graphical models with support for stochastic
branches, in the form of conditionals, loops, and recursion. Because
they are highly expressive, they pose a challenge in the design of
appropriate inference algorithms. This work focuses on Sequential
Monte Carlo (SMC) inference algorithms~\citep{DelMoral2006}, extending
an arc of research that includes probabilistic programming languages
(PPLs) such as Venture~\citep{Mansinghka2014}, Anglican~\citep{Tolpin2016},
Probabilistic C~\citep{Paige2014a}, WebPPL~\citep{Goodman2014},
Figaro~\citep{Pfeffer2016}, and Turing~\citep{Ge2016}, as well
as similarly-motivated software such as LibBi~\citep{Murray2015}
and BiiPS~\citep{Todeschini2014}.

The simplest SMC method, the bootstrap particle filter~\citep{Gordon1993},
requires only simulation—not pointwise evaluation—of the prior distribution.
While widely applicable, it may be suboptimal with respect to Monte
Carlo variance in situations where, in fact, pointwise evaluation
is possible, so that other options are viable. One way of reducing
Monte Carlo variance is to exploit analytical relationships between
random variables, such as conjugate priors and affine transformations.
Within SMC, this translates to improvements such as the locally-optimal
proposal, variable elimination, and Rao–Blackwellization (see \citep{Doucet2011}
for an overview). The present work seeks to automate such improvements
for the user of a PPL.

Typically, a probabilistic program must be run in order to discover
the relationships between random variables. Because of stochastic
branches, different runs may discover different relationships, or
even different random variables. While an equivalent graphical model
might be constructed for any single run, it would constitute only
partial observation. It may take many runs to observe the full model,
if this is possible in finite time at all. We therefore seek a runtime
mechanism for the solution of analytically-tractable substructure,
rather than a compile-time mechanism of static analysis.

A general-purpose programming language can be augmented with some
additional constructs, called \emph{checkpoints}, to produce a PPL
(see e.g. \citep{Tolpin2016}). Two checkpoints are usual, denoted
$\func{sample}$ and $\func{observe}$. The first suggests that a
value for a random variable needs to be sampled, the second that a
value for a random variable is given and needs to be conditioned upon.
At these checkpoints, random behavior may occur in the otherwise-deterministic
execution of the program, and intervention may be required by an inference
algorithm to produce a correct result.

The simplest inference algorithm instantiates a random variable when
first encountered at a $\func{sample}$ checkpoint, and updates a
weight with the likelihood of a given value at an $\func{observe}$
checkpoint. This produces samples from the prior distribution, weighted
by their likelihood under the observations. It corresponds to importance
sampling with the posterior as the target and the prior as the proposal.
A more sophisticated inference algorithm runs multiple instances of
the program simultaneously, pausing after each $\func{observe}$ checkpoint
to resample amongst executions. This corresponds to the bootstrap
particle filter (see e.g. \citep{Tolpin2016}).

These are \emph{forward} methods, in the sense that checkpoints are
executed in the order encountered, and sampling is myopic of future
observations. The present work introduces a mechanism to change the
order in which checkpoints are executed so that sampling can be informed
by future observations, exploiting analytical relationships between
random variables. This facilitates more sophisticated \emph{forward-backward}
methods, in the sense that information from future observations can
be propagated backward through the program.

We refer to this new mechanism as \emph{delayed sampling}. When a
$\func{sample}$ checkpoint is reached, its execution is delayed.
Instead, a new node representing the random variable is inserted into
a graph that is maintained alongside the running program. This graph
resembles a directed graphical model of those random variables encountered
so far that are involved in analytically-tractable relationships.
Each node of the graph is marginalized and conditioned by analytical
means for as long as possible until, eventually, it must be instantiated
for the program to continue execution. This occurs when the random
variable is passed as an argument to a function for which no analytical
overload is provided. It is at this last possible moment that sampling
is executed and the random variable instantiated.

Operations on the graph are forward-backward. The forward pass is
a filter, marginalizing each latent variable over its parents and
conditioning on observations, in all cases analytically. The backward
pass produces a joint sample. This has some similarity to belief propagation~\citep{Pearl1988},
but the backward passes differ: belief propagation typically obtains
the marginal posterior distribution of each variable, not a joint
sample. Furthermore, in delayed sampling the graph evolves dynamically
as the program executes, and at any time represents only a fraction
of the full model. This means that some heuristic decisions must be
made without complete knowledge of the model structure.

For SMC, delayed sampling yields locally-optimal proposals, variable
elimination, and Rao–Blackwellization, with some limitations, to be
detailed later. At worst, it provides no benefit. There is little
intrusion of the inference algorithm into modeling code, and possibly
no intrusion with appropriate language support. This is important,
as we consider the user experience and ergonomics of a PPL to be of
primary importance.

Related work has considered analytical solutions to probabilistic
programs. Where a full analytical solution is possible, it can be
achieved via symbolic manipulations in Hakaru~\citep{Shan2017}.
Where not, partial solutions using compile-time program transformations
are considered in \citep{Nori2014} to improve the acceptance rate
of Metropolis–Hastings algorithms. This compile-time approach requires
careful treatment of stochastic branches, and even then it may not
be possible to propagate analytical solutions through them. Delayed
sampling instead operates dynamically, at runtime. It handles stochastic
branches without problems, but may introduce some additional execution
overhead.

The paper is organized as follows. Section \ref{sec:methods} introduces
the delayed sampling mechanism. Section \ref{sec:experiments} provides
a set of pedagogical examples and two empirical case studies. Section
\ref{sec:discussion} discusses some limitations and future work.
Supplementary material includes further details of the case studies
and implementations.

\section{METHODS\label{sec:methods}}

As a probabilistic program runs, its memory state evolves dynamically
and stochastically over time, and can be considered a stochastic process.
Let $t=1,2,\ldots$ index a sequence of checkpoints. These checkpoints
may differ across program runs (this is one of the challenges of inference
for probabilistic programs, see e.g. \citep{Wingate2011}). In contrast
to the two-checkpoint $\func{sample}$-$\func{observe}$ formulation,
we define three checkpoint types:
\begin{itemize}
\item $\func{assume}(X,p(\cdot))$ to initialize a random variable $X$
with prior distribution $p(\cdot)$,
\item $\func{observe}(x,p(\cdot))$ to condition on a random variable $X$
with likelihood $p(\cdot)$ having some value $x$,
\item $\func{value}(X)$ to realize a value for a random variable $X$ previously
encountered at an $\func{assume}$ checkpoint.
\end{itemize}
We use the statistics convention that an uppercase character (e.g.
$X$) denotes a random variable, while the corresponding lowercase
character (e.g. $x$) denotes an instantiation of it.

An\emph{ $\func{assume}$} checkpoint does not result in a random
variable being sampled: its sampling is delayed until later. A \emph{$\func{value}$}
checkpoint occurs the first time that a random variable, previously
encountered by an \emph{$\func{assume}$}, is used in such a way that
its value is required. At this point it cannot be delayed any longer,
and is sampled.

Denote the state of the running program at checkpoint $t$ by $X_{t}\in\mathbb{X}_{t}$.
This can be interpreted as the current memory state of the program.
Randomness is exogenous and represented by the random process $U_{t}\in\mathbb{U}_{t}$.
This may be, for example, random entropy, a pseudorandom number sequence,
or uniformly distributed quasirandom numbers.

The program is a sequence of functions $f_{t}$ that each maps a starting
state $X_{t-1}=x_{t-1}$ and random input $U_{t}=u_{t}$ to an end
state $X_{t}=x_{t}$, so that $x_{t}=f_{t}(x_{t-1},u_{t})$. Note
that $f_{t}$ is a deterministic function given its arguments. It
is not permitted that $f_{t}$ has any intrinsic randomness, only
the extrinsic randomness provided by $U_{t}$.

The target distribution over $X_{t}$ is $\pi_{t}(\mathrm{d}x_{t})$,
typically a Bayesian posterior. In general, the program cannot sample
from this directly. Instead, it samples $x_{t}$ from some proposal
distribution $q_{t}(\mathrm{d}x_{t})$, which in many cases is just
the prior distribution $p_{t}(\mathrm{d}x_{t})$. Then, assuming that
both $\pi_{t}$ and $q_{t}$ admit densities, it computes an associated
importance weight $w_{t}\propto\pi_{t}(x_{t})/q_{t}(x_{t})$. Assuming
$U_{t}$ is distributed according to $\xi_{t}(\mathrm{d}u_{t})$,
we have
\[
q_{t}(\mathrm{d}x_{t})=\int_{\mathbb{X}_{t-1}}\int_{\mathbb{U}_{t}}\delta_{f_{t}(x_{t-1},u_{t})}(\mathrm{d}x_{t})\xi_{t}(\mathrm{d}u_{t})q_{t-1}(\mathrm{d}x_{t-1}),
\]
where $\delta$ is the Dirac measure. For brevity, we omit the subscript
$t$ henceforth, and simply update the state for the next time, as
though it is mutable.

\subsection{Motivation\label{sec:motivation}}

We are motivated by variance reduction in Monte Carlo estimators.
Consider some functional $\varphi(X)$ of interest. We wish to compute
expectations of the form:
\begin{align*}
\mathbb{E}_{\pi}[\varphi(X)] & =\int_{\mathbb{X}}\varphi(x)\pi(\mathrm{d}x)=\int_{\mathbb{X}}\varphi(x)\frac{\pi(x)}{q(x)}q(\mathrm{d}x).
\end{align*}
Self-normalized importance sampling estimates can be formed by running
the program $N$ times and computing (where superscript $n$ indicates
the $n$th program run):
\[
\hat{\varphi}:=\sum_{n=1}^{N}\bar{w}^{n}\varphi(x^{n}),\quad\bar{w}^{n}=w^{n}\bigg/\sum_{n=1}^{N}w^{n}.
\]
A classic aim is to reduce mean squared error:
\[
\mathrm{MSE}(\hat{\varphi})=\mathbb{E}_{q}\left[\left(\hat{\varphi}-\mathbb{E}_{\pi}[\varphi(X)]\right)^{2}\right].
\]
One technique to do so is \emph{Rao–Blackwellization}~(see e.g. \citep[§4.2]{Robert2004}).
Assume that, amongst the state $X$, there is some variable $X_{v}$
which has been observed to have value $x_{v}$, some set of variables
$X_{M}$ which can be marginalized out analytically, and some other
set of variables $X_{R}$ which have been instantiated previously.
The functional of interest is the incremental likelihood of $x_{v}$.
An estimator would usually require instantiation of $X_{M}^{n}\sim p(\mathrm{d}x_{M}^{n}\mid x_{R}^{n})$
for $n=1,\ldots,N$, and computation of:
\[
\hat{Z}:=\sum_{n=1}^{N}\bar{w}^{n}p(x_{v}\mid x_{M}^{n},x_{R}^{n}).
\]
The Rao–Blackwellized estimator does not instantiate $X_{M}$, but
rather marginalizes it out:
\[
\hat{Z}_{RB}:=\sum_{n=1}^{N}\bar{w}^{n}\int p(x_{v}\mid x_{M}^{n},x_{R}^{n})p(\mathrm{d}x_{M}^{n}\mid x_{R}^{n}).
\]
By the law of total variance, $\mathrm{var}(\hat{Z}_{RB})\leq\mathrm{var}(\hat{Z})$,
and as $\hat{Z}$ and $\hat{Z}_{RB}$ are unbiased~\citep{DelMoral2004},
$\mathrm{MSE}(\hat{Z}_{RB})\leq\mathrm{MSE}(\hat{Z})$.

This form of Rao–Blackwellization is local to each checkpoint. While
$X_{M}$ is marginalized out, it may require instantiation at future
checkpoints, and so it must also be possible to simulate $p(\mathrm{d}x_{M}\mid x_{v},x_{R})$.

\subsection{Delayed sampling\label{sec:details}}

Delayed sampling uses analytical relationships to reorder the execution
of checkpoints and reduce variance. Each $\func{observe}$ is executed
as early as possible, and the sampling associated with $\func{assume}$
is delayed for as long as possible, to be informed by observations
in between.

Alongside the state $X$, we maintain a graph $G=(V,E)$. This is
a directed graph consisting of a set of nodes $V$ and set of edges
$E\subset V\times V$, where $(u,v)\in E$ indicates a directed edge
from a parent node $u$ to a child node $v$. For $v\in V$, let $\Pa(v)=\{u\in V\mid(u,v)\in E\}$
denote its set of parents, and $\Ch(v)=\{u\in V\mid(v,u)\in E\}$
its set of children. Associated with each $v\in V$ is a random variable
$X_{v}$ (part of the state, $X$) and prior probability distribution
$p_{v}(\mathrm{d}x_{v}\mid x_{\Pa(v)})$, now using the subscript
of $X$ to select that part of the state associated with a single
node, or set of nodes. We partition $V$ into three disjoint sets
according to three states. Let
\begin{itemize}
\item $I\subseteq V$ be the set of nodes in an \emph{initialized} state,
\item $M\subseteq V$ be the set of nodes in a \emph{marginalized} state,
\item $R\subseteq V$ be the set of nodes in a \emph{realized} state.
\end{itemize}
At some checkpoint, the program would usually have instantiated all
variables in $V$ with a simulated or observed value, whereas under
delayed sampling only those in $R$ are instantiated, while those
in $I\cup M$ are delayed.

We will restrict the graph $G$ to be a forest of zero or more disjoint
trees, such that each node has at most one parent. This condition
is easily ensured by construction: the implementation makes anything
else impossible, i.e. only relationships between pairs of random variables
are coded. There are some interesting relationships that cannot be
represented as trees, such as a normal distribution with conjugate
prior over both mean and variance, or multivariate normal distributions.
We deal with these as special cases, collecting multiple nodes into
single supernodes and implementing relationships between pairs of
supernodes, much like the structure achieved by the junction tree
algorithm~\citep{Lauritzen1988}.

The following invariants are preserved at all times:
\begin{align}
\text{1. } & \text{If a node is in \emph{M} then its parent is in \emph{M}.}\label{eq:invariant-1}\\
\text{2. } & \text{A node has at most one child in \emph{M}.}\label{eq:invariant-2}
\end{align}

These imply that the nodes of $M$ form marginalized paths: one in
each of the disjoint trees of $G$, from the root node to a node (possibly
itself) in the same tree. We will refer to the unique such path in
each tree as its \emph{$M$-path}. The node at the start of the \emph{$M$}-path
is a root node, while the node at the end is referred to as a \emph{terminal}
node. Terminal nodes have a special place in the algorithms below,
and are denoted by the set $T$.

By the invariants, each $v\in M\setminus T$ has a child $u\in M$;
let $\Fo(v)$ denote the entire subtree with this child $u$ as its
root (the \emph{forward} set). Otherwise let $\Fo(v)$ be the empty
set. The graph $G$ then encodes the distribution
\begin{align}
 & \left(\prod_{v\in I}q_{v}(\mathrm{d}x_{v}\mid x_{\Pa(v)})\right)\left(\prod_{v\in M\setminus T}q_{v}(\mathrm{d}x_{v}\mid x_{R\setminus\Fo(v)})\right)\times\nonumber \\
 & \quad\left(\prod_{v\in T}q_{v}(\mathrm{d}x_{v}\mid x_{R})\right),\label{eq:graph-invariant}
\end{align}
where $q_{v}$ equals the prior for nodes in $I$, some updated distribution
for nodes in $M$, and all nodes in $R$ are instantiated. The distribution
suggests why terminals (in the set $T$) are important: they are the
nodes informed by all instantiated random variables up to the current
point in the program, and can be immediately instantiated themselves.
Other nodes in \emph{$M$} await information to be propagated backward
from their forward set before they, too, can be instantiated.

\begin{figure*}[t]
\resizebox{\textwidth}{!}{\begin{tikzpicture}
\node (a) at (0, 0) [m-node] {$a$};
\node (b) at (0, -1.5) [m-node] {$b$};
\node (c) at (-0.5, -3) [t-node] {$c$};
\node (d) at (0.6, -3) [i-node] {$d$};
\node (e) at (-1, -4.5) [i-node] {$e$};
\node (f) at (0, -4.5) [i-node] {$f$};

\node [right=0.2cm of a,align=left] {$a \sim \N(0, 1)$\\ $q_a = \N(0, 1)$};
\node [right=0.2cm of b,align=left] {$b \sim \N(a, 1)$\\ $q_b = \N(0, 2)$};
\node [left=0.1cm of c,align=right] {$c \sim \N(b, 1)$\\ $q_c = \N(0, 3)$};
\node [right=0.2cm of d,align=left] {$d \sim \N(b, 1)$};
\node [left=0.1cm of e,align=right] {$e \sim \N(c, 1)$};
\node [right=0.2cm of f,align=left] {$f \sim \N(c, 1)$};

\draw [stem] (a) -- (b);
\draw [stem] (b) -- (c);
\draw [edge] (b) -- (d);
\draw [edge] (c) -- (e);
\draw [edge] (c) -- (f);

\node (a2) at (11, 0) [m-node] {$a$};
\node (b2) at (11, -1.5) [m-node] {$b$};
\node (c2) at (10.5, -3) [r-node] {$c$};
\node (d2) at (11.5, -3) [t-node] {$d$};
\node (e2) at (10, -4.5) [m-node] {$e$};
\node (f2) at (11, -4.5) [m-node] {$f$};

\node [right=0.3cm of a2,align=left] {$a \sim \N(0, 1)$\\ $q_a = \N(0, 1)$};
\node [right=0.3cm of b2,align=left] {$b \sim \N(a, 1)$\\ $\color{altblue} q_b = \N(\frac{4}{3}, \frac{2}{3})$};
\node [left=0.3cm of c2,align=right] {$c \sim \N(b, 1)$\\ $\color{altblue} c = 2$\\(sampled)};
\node [right=0.5cm of d2,align=left] {$d \sim \N(b, 1)$\\ $\color{altblue} q_d = \N(\frac{4}{3}, \frac{5}{3})$};
\node [left=0.2cm of e2,align=right] {$e \sim \N(c, 1)$\\ $\color{altblue} q_e = \N(2, 1)$};
\node [right=0.3cm of f2,align=left] {$f \sim \N(c, 1)$\\ $\color{altblue} q_f = \N(2, 1)$};

\draw [stem] (a2) -- (b2);
\draw [stem] (b2) -- (d2);

\node [align=left] at (5.5, -1.25) {
\sc \color{altblue}Graft($d$)\\
\sc \hspace{0.2cm}Graft($b$)\\
\sc \hspace{0.4cm}Prune($c$)\\
\sc \hspace{0.6cm}Sample($c$)\\
\sc \hspace{0.8cm}Realize($c$)\\
\sc \hspace{1cm}Marginalize($e$)\\
\sc \hspace{1cm}Marginalize($f$)\\
\sc \hspace{0.2cm}Marginalize($d$)};
\draw [->,ultra thick](3.5,-3) -- (7.5,-3);
\end{tikzpicture}}

\caption{Demonstration of the \emph{$M$}-path and operations on the graph.
On the left, the \emph{$M$}-path reaches from the root node, $a$,
to the terminal node, $c$, marked in bold lines. The $\proc{Graft}$
operation is called for $d$. This requires a realignment of the \emph{$M$}-path
around $b$, pruning the previous \emph{$M$}-path at $c$, then extending
it through to $d$. The stack trace of operations is in the center,
and the final state on the right. Descendants of $c$ that were not
on the \emph{$M$}-path are now the roots of separate, disjoint trees.
\label{fig:realign}}
\end{figure*}
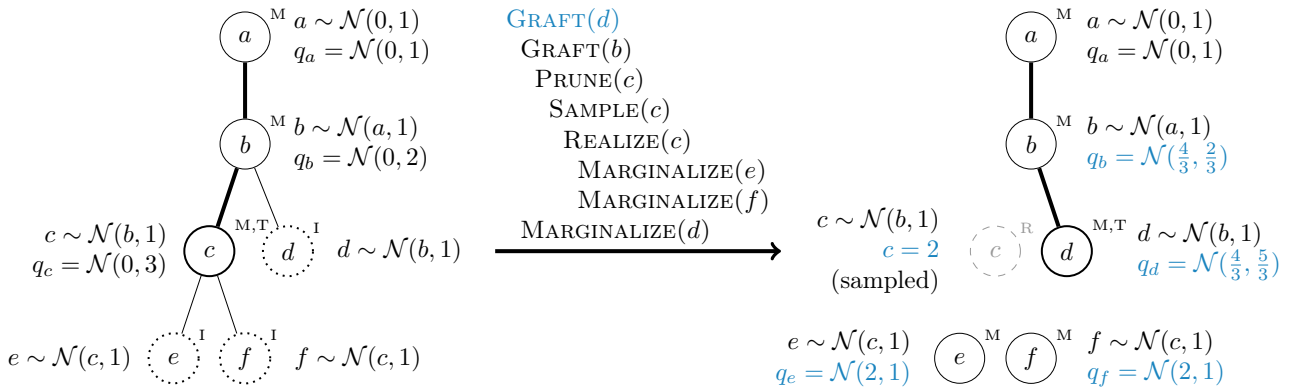

When the program reaches a checkpoint, it triggers operations on the
graph (details follow):
\begin{itemize}
\item For $\func{assume}(X_{v},p(\cdot)$), call $\proc{Initialize}(v,p(\cdot))$,
which inserts a new node $v$ into the graph.
\item For $\func{observe}(x_{v},p(\cdot))$, call $\proc{Initialize}(v,p(\cdot))$,
then $\proc{Graft}(v)$, which turns $v$ into a terminal node, then
$\proc{Observe}(v)$, which assigns the observed value to $v$ and
updates its parent by conditioning.
\item For $\func{value}(X_{v})$, call $\proc{Graft}(v)$, then $\proc{Sample}(v)$,
which samples a value for $v$.
\end{itemize}
Figure~\ref{fig:operations} provides pseudocode for all operations;
Figure~\ref{fig:realign} illustrates their combination. Operations
are of two types: \emph{local} and \emph{recursive}. Local operations
modify a single node and possibly its parent:
\begin{itemize}
\item $\proc{Initialize}(v,p(\cdot))$ inserts a new node $v$ into the
graph. If $v$ requires a parent, $u$ (implied by $p(\cdot)$ having
a conditional form, i.e. $p(\mathrm{d}x_{v}\mid x_{u})$ not $p(\mathrm{d}x_{v}$)),
then $v$ is put in $I$ and the edge $(u,v)$ inserted. Otherwise,
it is a root node and is put in $M$, with no edges inserted.
\item $\proc{Marginalize}(v)$, where $v$ is the child of a terminal node,
moves $v$ from $I$ to $M$ and updates its distribution by marginalizing
over its parent.
\item $\proc{Sample}(v)$ or $\proc{Observe}(v)$, where $v$ is a terminal
node, assigns a value to the associated random variable by either
sampling or observing, moves $v$ from $M$ to $R$, and updates the
distribution of its parent node by conditioning. Both $\proc{Sample}(v)$
and $\proc{Observe}(v)$ use an auxiliary function $\proc{Realize}(v)$
for their common operations.
\end{itemize}
As shown in the pseudocode, these local operations have strict preconditions
that limit their use to only a subset of the nodes of the graph, e.g.
only terminal nodes may be sampled or observed. As long as these preconditions
are satisfied, the invariants (\ref{eq:invariant-1}) and (\ref{eq:invariant-2})
are maintained, and the graph $G$ encodes the representation (\ref{eq:graph-invariant}).
This is straightforward to check.

The recursive operations realign the $M$-path to establish the preconditions
for any given node, so that local operations may be applied to it.
These have side effects, in that other nodes may be modified to achieve
the realignment. The key recursive operation is $\proc{Graft}$, which
combines local operations to extend the $M$-path to a given node,
making it a terminal node. Internally, $\proc{Graft}$ may call another
recursive operation, $\proc{Prune}$, to shorten the existing $M$-path
by realizing one or more variables.

\begin{figure}
\input{initialise.tex}

\input{marginalise.tex}

\input{sample.tex}

\input{observe.tex}

\input{realise.tex}

\input{graft.tex}

\input{prune.tex}

\caption{Operations on the graph. The left arrow ($\leftarrow$) denotes assignment.
Assigning to a distribution is interpreted as updating its hyperparameters.
\label{fig:operations}}
\end{figure}

\begin{table*}[p]
\begin{tabular}{lll>{\raggedright}p{0.44\textwidth}}
\noalign{\vskip0.5mm}
\textbf{\footnotesize{}Program} & \textbf{\footnotesize{}Checkpoint} & \multicolumn{1}{l}{\textbf{\footnotesize{}Local operations}} & \textbf{\footnotesize{}Commentary}\tabularnewline[0.5mm]
\hline
\noalign{\vskip0.5mm}
\texttt{\scriptsize{}x \texttildemiddle~N(0,1);} & {\footnotesize{}$\func{assume}(X)$} & {\footnotesize{}$\proc{Initialize}(X)$} & \multirow{2}{0.44\textwidth}{{\footnotesize{}Named }\texttt{\footnotesize{}delay\_triplet}{\footnotesize{}
in supplementary material.}}\tabularnewline[0.5mm]
\noalign{\vskip0.5mm}
\texttt{\scriptsize{}y \texttildemiddle~N(x,1);} & {\footnotesize{}$\func{assume}(Y)$} & {\footnotesize{}$\proc{Initialize}(Y)$} & \tabularnewline[0.5mm]
\noalign{\vskip0.5mm}
\texttt{\scriptsize{}\uline{z}}\texttt{\scriptsize{} \texttildemiddle~N(y,1);} & {\footnotesize{}$\func{observe}(z)$} & {\footnotesize{}$\proc{Initialize}(Z)$} & \tabularnewline[0.5mm]
\noalign{\vskip0.5mm}
 &  & {\footnotesize{}$\proc{Marginalize}(Y)$} & \multirow{3}{0.44\textwidth}{{\footnotesize{}No $\proc{Marginalize}(X)$ is necessary: $X$, as
a root node, is initialized in the marginalized state.}}\tabularnewline[0.5mm]
\noalign{\vskip0.5mm}
 &  & {\footnotesize{}$\proc{Marginalize}(Z)$} & \tabularnewline[0.5mm]
\noalign{\vskip0.5mm}
 &  & {\footnotesize{}$\proc{Observe}(z)$} & \tabularnewline[0.5mm]
\noalign{\vskip0.5mm}
\texttt{\scriptsize{}print(x);} & {\footnotesize{}$\func{value}(X)$} & {\footnotesize{}$\proc{Sample}(Y)$} & {\footnotesize{}Samples $Y\sim p(\mathrm{d}y\mid z)$.}\tabularnewline[0.5mm]
\noalign{\vskip0.5mm}
 &  & {\footnotesize{}$\proc{Sample}(X)$} & {\footnotesize{}Samples $X\sim p(\mathrm{d}x\mid y,z)$.}\tabularnewline[0.5mm]
\noalign{\vskip0.5mm}
\texttt{\scriptsize{}print(y);} &  &  & {\footnotesize{}A value $Y=y$ is already known.}\tabularnewline[0.5mm]
\hline
\noalign{\vskip0.5mm}
\texttt{\scriptsize{}x \texttildemiddle~N(0,1);} & {\footnotesize{}$\func{assume}(X)$} & {\footnotesize{}$\proc{Initialize}(X)$} & \multirow{5}{0.44\textwidth}{{\footnotesize{}Named }\texttt{\footnotesize{}delay\_iid}{\footnotesize{}
in supplementary material. It encodes multiple i.i.d. observations
with a conjugate prior distribution over their mean.}}\tabularnewline[0.5mm]
\noalign{\vskip0.5mm}
\texttt{\scriptsize{}for (t in 1..T) \{} &  &  & \tabularnewline[0.5mm]
\noalign{\vskip0.5mm}
\texttt{\scriptsize{}~~}\texttt{\scriptsize{}\uline{y{[}t{]}}}\texttt{\scriptsize{}
\texttildemiddle~N(x,1);} & {\footnotesize{}$\func{observe}(y_{t})$} & {\footnotesize{}$\proc{Initialize}(y_{t})$} & \tabularnewline[0.5mm]
\noalign{\vskip0.5mm}
 &  & {\footnotesize{}$\proc{Marginalize}(y_{t})$} & \tabularnewline[0.5mm]
\noalign{\vskip0.5mm}
 &  & {\footnotesize{}$\proc{Observe}(y_{t})$} & \tabularnewline[0.5mm]
\noalign{\vskip0.5mm}
{\scriptsize{}\}} &  &  & \tabularnewline[0.5mm]
\noalign{\vskip0.5mm}
\texttt{\scriptsize{}print(x);} & {\footnotesize{}$\func{value}(X)$} & {\footnotesize{}$\proc{Sample}(X)$} & {\footnotesize{}Samples $X\sim p(\mathrm{d}x\mid y_{1},\ldots,y_{T})$.}\tabularnewline[0.5mm]
\hline
\noalign{\vskip0.5mm}
\texttt{\scriptsize{}x \texttildemiddle~Bernoulli(p);} & {\footnotesize{}$\func{assume}(X)$} & {\footnotesize{}$\proc{Initialize}(X)$} & \multirow{2}{0.44\textwidth}{{\footnotesize{}Named }\texttt{\footnotesize{}delay\_spike\_and\_slab}{\footnotesize{}
in supplementary material. It encodes a spike-and-slab prior~\citep{Mitchell1988}
often used in Bayesian linear regression.}}\tabularnewline[0.5mm]
\noalign{\vskip0.5mm}
\texttt{\scriptsize{}if (x) \{} & {\footnotesize{}$\func{value}(X)$} & {\footnotesize{}$\proc{Sample}(X)$} & \tabularnewline[0.5mm]
\noalign{\vskip0.5mm}
\texttt{\scriptsize{}~~y \texttildemiddle~N(0,1);} & {\footnotesize{}$\func{assume}(Y)$} & {\footnotesize{}$\proc{Initialize}(Y)$} & \tabularnewline[0.5mm]
\noalign{\vskip0.5mm}
\texttt{\scriptsize{}\} else \{} &  &  & \tabularnewline[0.5mm]
\noalign{\vskip0.5mm}
\texttt{\scriptsize{}~~y <- 0;} &  &  & {\footnotesize{}Used as a regular variable, no graph operations are
triggered.}\tabularnewline[0.5mm]
\noalign{\vskip0.5mm}
\texttt{\scriptsize{}\}} &  &  & {\footnotesize{}$Y$ is marginalized or realized as some $Y=y$ by
the end, according to the stochastic branch.}\tabularnewline[0.5mm]
\hline
\noalign{\vskip0.5mm}
\texttt{\scriptsize{}x{[}1{]} \texttildemiddle~N(0,1);} & {\footnotesize{}$\func{assume}(X_{1})$} & {\footnotesize{}$\proc{Initialize}(X_{1})$} & \multirow{3}{0.44\textwidth}{{\footnotesize{}Named }\texttt{\footnotesize{}delay\_kalman}{\footnotesize{}
in supplementary material. It encodes a linear-Gaussian state-space
model, for which delayed sampling yields a forward Kalman filter and
backward simulation.}}\tabularnewline[0.5mm]
\noalign{\vskip0.5mm}
\texttt{\scriptsize{}\uline{y{[}1{]}}}\texttt{\scriptsize{} \texttildemiddle~N(x{[}1{]},1);} & {\footnotesize{}$\func{observe}(y_{1})$} & {\footnotesize{}$\proc{Initialize}(y_{1})$} & \tabularnewline[0.5mm]
\noalign{\vskip0.5mm}
 &  & {\footnotesize{}$\proc{Marginalize}(y_{1})$} & \tabularnewline[0.5mm]
\noalign{\vskip0.5mm}
 &  & {\footnotesize{}$\proc{Observe}(y_{1})$} & \tabularnewline[0.5mm]
\noalign{\vskip0.5mm}
\texttt{\scriptsize{}for (t in 2..T) \{} &  &  & \multirow{7}{0.44\textwidth}{{\footnotesize{}After each $t$th iteration of this loop, the distribution
$p(\mathrm{d}x_{t}\mid y_{1},\ldots,y_{t})$ is obtained; the behavior
corresponds to a Kalman filter.}}\tabularnewline[0.5mm]
\noalign{\vskip0.5mm}
\texttt{\scriptsize{}~~x{[}t{]} \texttildemiddle~N(a{*}x{[}t-1{]},1);} & {\footnotesize{}$\func{assume}(X_{t})$} & {\footnotesize{}$\proc{Initialize}(X_{t})$} & \tabularnewline[0.5mm]
\noalign{\vskip0.5mm}
\texttt{\scriptsize{}~~}\texttt{\scriptsize{}\uline{y{[}t{]}}}\texttt{\scriptsize{}
\texttildemiddle~N(x{[}t{]},1);} & {\footnotesize{}$\func{observe}(y_{t})$} & {\footnotesize{}$\proc{Initialize}(y_{t})$} & \tabularnewline[0.5mm]
\noalign{\vskip0.5mm}
 &  & {\footnotesize{}$\proc{Marginalize}(X_{t})$} & \tabularnewline[0.5mm]
\noalign{\vskip0.5mm}
 &  & {\footnotesize{}$\proc{Marginalize}(y_{t})$} & \tabularnewline[0.5mm]
\noalign{\vskip0.5mm}
 &  & {\footnotesize{}$\proc{Observe}(y_{t})$} & \tabularnewline[0.5mm]
\noalign{\vskip0.5mm}
{\scriptsize{}\}} &  &  & \tabularnewline[0.5mm]
\noalign{\vskip0.5mm}
\texttt{\scriptsize{}print(x{[}1{]});} & {\footnotesize{}$\func{value}(X_{1})$} & {\footnotesize{}$\proc{Sample}(X_{T})$} & {\footnotesize{}Samples $X_{T}\sim p(\mathrm{d}x_{T}\mid y_{1},\ldots,y_{T})$.}\tabularnewline[0.5mm]
\noalign{\vskip0.5mm}
 &  & {\footnotesize{}$\ldots$} & {\footnotesize{}Recursively samples $X_{t}\sim p(\mathrm{d}x_{t}\mid x_{t+1},y_{1},\ldots,y_{t})$
and computes $p(\mathrm{d}x_{t-1}\mid x_{t},y_{1},\ldots,y_{t-1})$.}\tabularnewline[0.5mm]
\noalign{\vskip0.5mm}
 &  & {\footnotesize{}$\proc{Sample}(X_{1})$} & {\footnotesize{}Samples $X_{1}\sim p(\mathrm{d}x_{1}\mid x_{2},y_{1})$.}\tabularnewline[0.5mm]
\hline
\end{tabular}

\caption{Pedagogical examples of delayed sampling applied to four probabilistic
programs, showing the programs themselves (first column), the checkpoints
reached as they execute linearly from top to bottom (second column),
the sequence of local operations that these trigger on the graph (third
column), and commentary (fourth column). The programs use a Birch-like
syntax. Random variables with given values (from earlier assignment)
are annotated by underlining. The function \texttt{print} is assumed
to accept real-valued arguments only, so may trigger a $\func{value}$
checkpoint when used.\label{tab:pedagogical-examples}}
\end{table*}

\section{EXAMPLES\label{sec:experiments}}

We have implemented delayed sampling in Anglican (see also \citep{Lunden2017})
and a new PPL called Birch. Details are given in Appendices \ref{app:anglican}
and \ref{app:birch}.

Table~\ref{tab:pedagogical-examples} provides pedagogical examples
using a Birch-like syntax, showing the sequence of checkpoints and
graph operations triggered as some simple programs execute. They show
how delayed sampling behaves through programming structures such as
conditionals and loops, including stochastic branches.

In addition, we provide two case studies where delayed sampling improves
inference, firstly a linear-nonlinear state-space model with simulated
data, secondly a vector-borne disease model with real data from an
outbreak of dengue virus in Micronesia. We use a simple random-weight
or pseudo-marginal-type importance sampling algorithm for both of
these examples:
\begin{enumerate}
\item Run SMC on the probabilistic program with delayed sampling enabled,
producing $N$ number of samples $x^{1},\ldots,x^{N}$ with associated
weights $w^{1},\ldots,w^{N}$ and a marginal likelihood estimate $\hat{Z}$.
\item Draw $a\in\{1,\ldots,N\}$ from the categorical distribution defined
by $P(a)=w^{a}/\sum_{n=1}^{N}w^{n}$.
\item Output $x^{a}$ with weight $\hat{Z}$.
\end{enumerate}
This produces one sample with associated weight, but may be repeated
as many times as necessary—in parallel, even—to produce an importance
sample as large as desired. The success of the approach depends on
the variance of $\hat{Z}$. This variance can be reduced by marginalizing
out one or more variables (recall Section \ref{sec:motivation}).
This is what delayed sampling achieves, and so we compare the variance
of $\hat{Z}$ with delayed sampling enabled and disabled. When disabled,
the SMC algorithm is simply a bootstrap particle filter. When enabled,
it yields a Rao–Blackwellized particle filter. Where parameters are
involved (as in the second case study), the diversity of parameter
values depletes through the resampling step of SMC. This has motivated
more sophisticated methods for parameter estimation such as particle
Markov chain Monte Carlo methods~\citep{Andrieu2010}, also applied
to probabilistic programs~\citep{Wood2014}. Particle Gibbs is an
obvious candidate here. We find, however, that the reduction in variance
afforded by marginalizing out one or more variables with delayed sampling
is sufficient to enable the above importance sampling algorithm for
the two case studies here.

\subsection{Linear-nonlinear state-space model\label{sec:linear-nonlinear-case-study}}

The first example is that of a mixed linear-nonlinear state-space
model. For this model, delayed sampling yields a particle filter with
locally-optimal proposal and Rao–Blackwellization.

The model is given by~\citep{Lindsten2010} and repeated in Appendix
\ref{sec:appendix-linear-nonlinear}. It consists of both nonlinear
and linear-Gaussian state variables, as well as nonlinear and linear-Gaussian
observations. Parameters are fixed. Ideally, the linear-Gaussian substructure
is solved analytically (e.g. using a Kalman filter), leaving only
the nonlinear substructure to sample (e.g. using a particle filter).
The Rao–Blackwellized particle filter, also known as the marginalized
particle filter, was designed to achieve precisely this~\citep{Chen2000,Schon2005}.

Delayed sampling automatically yields this method for this model,
as long as analytical relationships between multivariate Gaussian
distributions are encoded. In Birch these are implemented as supernodes:
single nodes in the graph that contain multiple random variables.
While the relationships between individual variables in a multivariate
Gaussian have, in general, directed acyclic graph structure, their
implementation as supernodes maintains the required tree structure.

\begin{figure*}[tp]
\begin{minipage}[t]{0.5\textwidth}%
\includegraphics[width=1\textwidth]{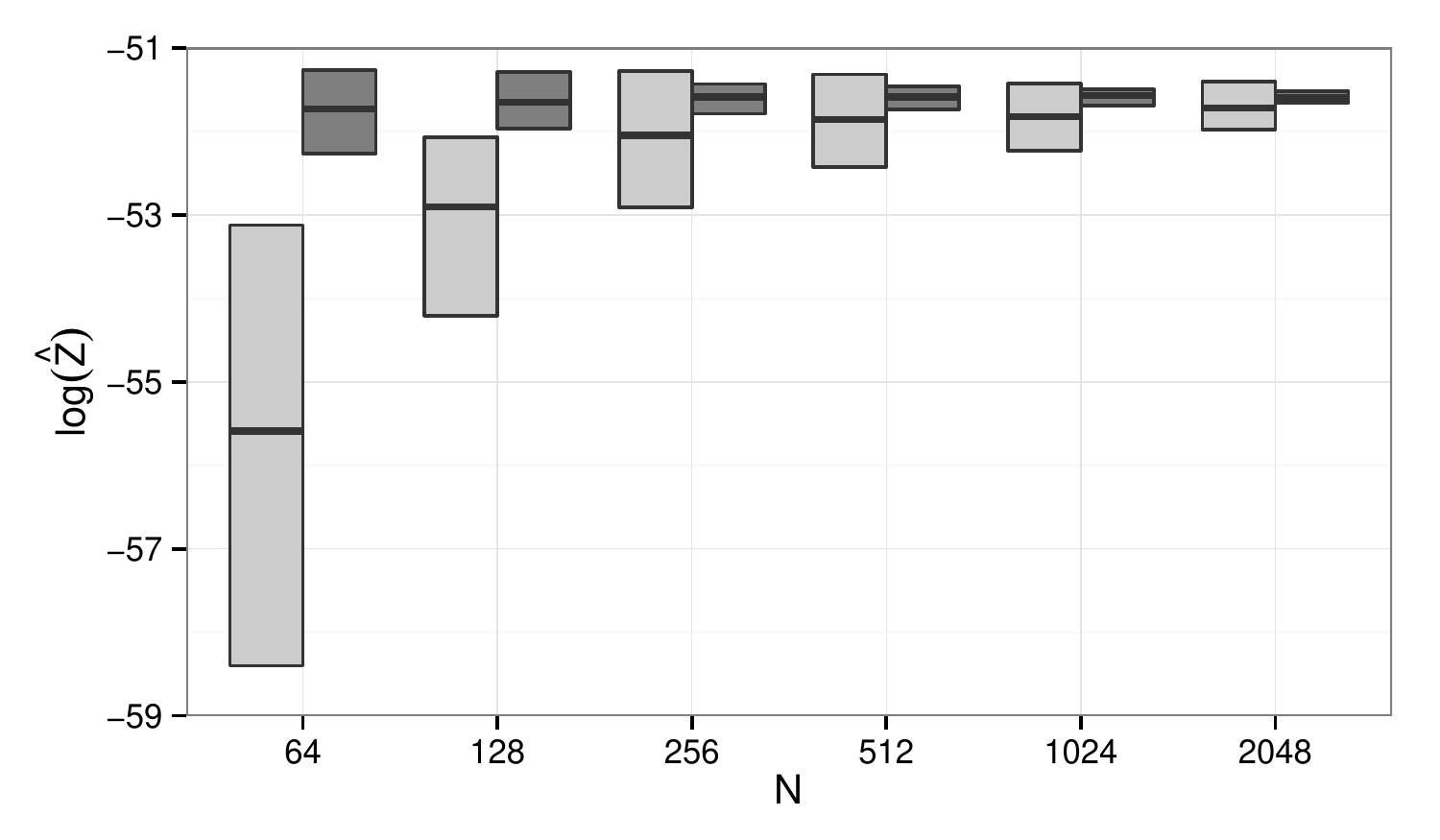}%
\end{minipage}%
\begin{minipage}[t]{0.5\textwidth}%
\includegraphics[width=1\textwidth]{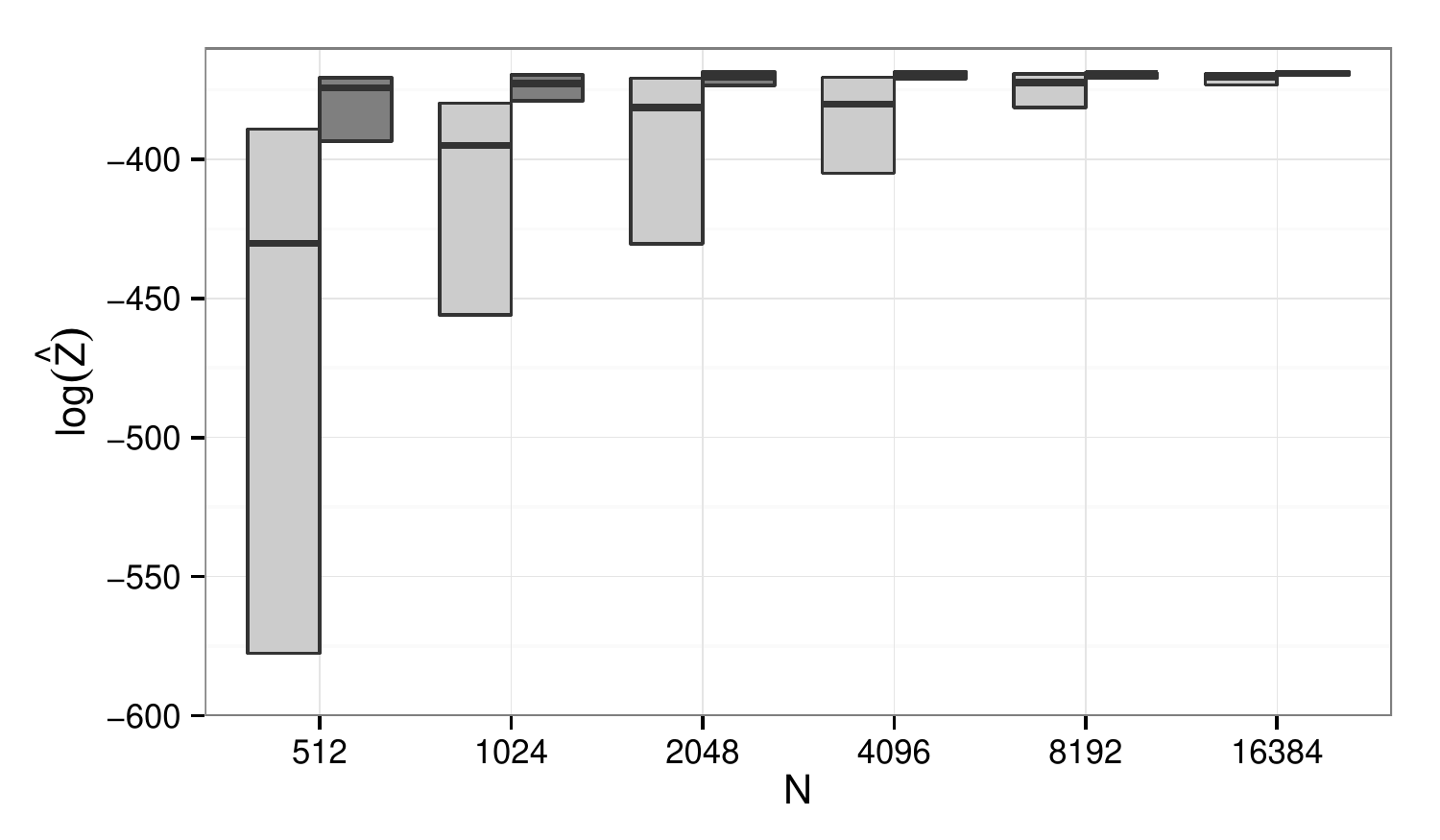}%
\end{minipage}

\caption{Distribution of the marginal log-likelihood estimate ($\log\hat{Z}$)
for different numbers of particles ($N$) over 100 runs for (left)
the linear-nonlinear state-space model, and (right) the vector-borne
disease model, with (light gray) delayed sampling disabled, corresponding
to a bootstrap particle filter, and (dark gray) delayed sampling enabled,
corresponding to a Rao–Blackwellized particle filter. All runs use
systematic resampling~\citep{Kitagawa1996} when effective sample
size falls below 0.7$N$. Boxes indicate the interquartile range,
midline the median. In both cases, significantly fewer particles are
required to achieve comparable variance when delayed sampling is enabled.\label{fig:rbpf}}
\end{figure*}

The model is run for 100 time steps to simulate data. It is run again
with SMC, conditioning on this data. For various numbers of particles,
it is run 100 times to estimate $\hat{Z}$, with delayed sampling
enabled and disabled. Figure \ref{fig:rbpf} (left) plots the distribution
of these estimates. Clearly, with delayed sampling enabled, fewer
particles are needed to achieve comparable variance in the log-likelihood
estimate.

\subsection{Vector-borne disease model\label{sec:vbd-cast-study}}

The second example is an epidemiological case study of an outbreak
of dengue virus: a mosquito-borne tropical disease with an estimated
50-100 million cases and 10000 deaths worldwide each year~\citep{Stanaway2016}.
It is based on the study in \citep{Funk2016a}, which jointly models
two outbreaks of dengue virus and one of Zika virus in two separate
locations (and populations) in Micronesia. Presented here is a simpler
study limited to one of those outbreaks, specifically that of dengue
on the Yap Main Islands in 2011. The data used consists of 172 observations
of reported cases, on a daily basis during the main outbreak, and
on a weekly basis before and after.

The model consists of two components, representing the human and mosquito
populations, coupled via cross-infection. Each population is further
divided into subpopulations of susceptible, exposed, infectious and
recovered individuals. At each time step a binomial transfer occurs
between subpopulations, parameterized with conjugate beta priors.
Details are in Appendix \ref{sec:appendix-vbd}.

The task is both parameter and state estimation. For this model, delayed
sampling produces a Rao–Blackwellized particle filter where parameters,
rather than state variables, are marginalized out. While the state
variables are sampled immediately, the parameters are maintained in
a marginalized state, conditioned on the samples of these state variables.
This is a consequence of conjugacy between the beta priors on parameters
and the binomial likelihoods of the state variables (as pseudo-observations).

For various numbers of particles, SMC is run 100 times to estimate
$\hat{Z}$, with delayed sampling enabled and disabled. Figure \ref{fig:rbpf}
(right) plots the distribution of these estimates. Clearly, with delayed
sampling enabled, fewer particles are needed to achieve comparable
variance in the log-likelihood estimate. Some posterior results are
given in Appendix \ref{sec:appendix-vbd}.

\section{DISCUSSION AND CONCLUSION\label{sec:discussion}}

Table~\ref{tab:pedagogical-examples} demonstrates how delayed sampling
operates through typical program structures such as conditionals and
loops, including stochastic branches as encountered in probabilistic
programs. Figure~\ref{fig:rbpf} demonstrates the potential gains.
These are particularly encouraging given that the mechanism is mostly
automatic.

Some limitations are worth noting. The graph of analytically-tractable
relationships must be a forest of disjoint trees. It is unclear whether
this is a significant limitation in practice, but support for more
general structures may be desirable. It is worth emphasizing that
this relates to the structure of analytically-tractable relationships
and the ability of the mechanism to utilize them, not to the structure
of the model as a whole. At present, for more general structures,
some opportunities for variance reduction are missed. One remedy is
to encode supernodes, as for the multivariate Gaussian distributions
in Section~\ref{sec:linear-nonlinear-case-study}.

Delayed sampling potentially reorders the sampling associated with
$\func{assume}$ checkpoints, and the interleaving of this amongst
$\func{observe}$ checkpoints, but does not reorder the execution
of $\func{observe}$ checkpoints. There is an opportunity cost to
this. Consider the final example in Table~\ref{tab:pedagogical-examples}:
move the observations\texttt{ $y_{1},\ldots,y_{T}$} into a second
loop that traverses time backward from $T$ to $1$. Delayed sampling
now draws each $x_{t}$ from $p(\mathrm{d}x_{t}\mid x_{t+1},y_{t})$,
not $p(\mathrm{d}x_{t}\mid x_{t+1},y_{1},\ldots,y_{t})$. This is
suboptimal but not incorrect: whatever the distribution, importance
weights correct for its discrepancy from the target. It is again unclear
whether this is a significant limitation in practice; examples seem
contrived and easily fixed by reordering code.

While delayed sampling may reduce the number of samples required for
comparable variance, it does require additional computation per sample.
For univariate relationships (e.g. beta-binomial, gamma-Poisson),
this overhead is constant and—we conjecture—likely worthwhile for
any fixed computational budget. For multivariate relationships the
overhead is more complex and may not be worthwhile (e.g. multivariate
Gaussian conjugacies require matrix inversions that are $\mathcal{O}(N^{3})$
in the number of dimensions). A thorough empirical comparison is beyond
the scope of this article.

Finally, while the focus of this work is SMC, delayed sampling may
be useful in other contexts. With undirected graphical models, for
example, delayed sampling may produce a collapsed Gibbs sampler. This
is left to future work.

\section*{Acknowledgements}

This research was financially supported by the Swedish Foundation
for Strategic Research (SSF) via the project \emph{ASSEMBLE}. Jan
Kudlicka was supported by the Swedish Research Council grant 2013-4853.

\section*{Supplementary material}

Appendix \ref{sec:appendix-linear-nonlinear} details the linear-nonlinear
state-space model, and Appendix \ref{sec:appendix-vbd} the vector-borne
disease model. Appendix \ref{app:anglican} details the Anglican implementation,
and Appendix \ref{app:birch} the Birch implementation. Code is included
for the pedagogical examples in both Anglican and Birch, and for the
empirical case studies, along with data sets, in Birch only.

\bibliographystyle{abbrvnat}
\bibliography{delay}

\clearpage{}

\appendix

\section{Details of the linear-nonlinear state-space model\label{sec:appendix-linear-nonlinear}}

The full model is described in \citep{Lindsten2010}. The state model
contains both nonlinear ($X_{t}^{n}$) and linear-Gaussian ($X_{t}^{l}$)
state variables, and is given by:
\begin{align*}
X_{0}^{n} & \sim\mathcal{N}(0,1)\\
X_{t}^{n} & \sim\mathcal{N}(\arctan x_{t-1}^{n}+Bx_{t-1}^{l},0.01)\\
X_{0}^{l} & \sim\mathcal{N}(0,I_{3\times3})\\
X_{t}^{l} & \sim\mathcal{N}(Ax_{t-1}^{l},0.01I_{3\times3}).
\end{align*}
The observation model contains both nonlinear ($Y_{t}^{n}$) and linear-Gaussian
($Y_{t}^{l}$) observations, and is given by:
\begin{align*}
Y_{t}^{n} & \sim\mathcal{N}\left(0.1(x_{t}^{n})^{2}\mathrm{sgn}(x_{t}^{n}),0.1\right)\\
Y_{t}^{l} & \sim\mathcal{N}(Cx_{t}^{l},0.1I_{3\times3}).
\end{align*}
Parameters are fixed as follows:
\begin{align*}
A & =\begin{pmatrix}1 & 0.3 & 0\\
0 & 0.92 & -0.3\\
0 & 0.3 & 0.92
\end{pmatrix}\\
B & =\begin{pmatrix}1 & 0 & 0\end{pmatrix}\\
C & =\begin{pmatrix}1 & -1 & 1\end{pmatrix}.
\end{align*}

\section{Details of the vector-borne disease model\label{sec:appendix-vbd}}

The process model is a discrete-time and discrete-state stochastic
model based on the continuous-time and continuous-state deterministic
mean-field approximation used in \citep{Funk2016a}. It consists of
two SEIR (susceptible, exposed, infectious, recovered) compartmental
models, one for the human population, the other for the mosquito population,
coupled via cross-infection terms. Each component consists of state
variables giving population counts in each of the four compartments:
$s$ (susceptible), $e$ (exposed), $i$ (infectious), and $r$ (recovered),
along with a total population $n$ that maintains the identity $n=s+e+i+r$,
and parameters $\nu$ (birth probability), $\mu$ (death probability),
$\lambda$ (transmission probability), $\delta$ (infectious probability),
and $\gamma$ (recovery probability). A susceptible human may become
infected when bitten by an infectious mosquito, while a susceptible
mosquito may become infected when biting an infectious human.

We use superscript $h$ to denote state variables and parameters associated
with the human component, and superscript $m$ to denote those associated
with the mosquito component. For state variables, subscripts index
time in days.

\subsection{Initial condition model}

For the setting of Yap Main Islands in 2011, the following initial
conditions are prescribed:
\begin{align*}
n_{0}^{h} & =7370 & n_{0}^{m} & =10^{u}n_{0}^{h}\\
s_{0}^{h} & =n_{0}^{h}-e_{0}^{h}-i_{0}^{h}-r_{0}^{h} & s_{0}^{m} & =n_{0}^{m}\\
e_{0}^{h} & \sim\mathrm{Poisson}(10) & e_{0}^{m} & =0\\
i_{0}^{h}-1 & \sim\mathrm{Poisson}(10) & i_{0}^{m} & =0\\
r_{0}^{h} & \sim\mathrm{Binomial}(n_{1}^{h},6/100) & r_{0}^{m} & =0,
\end{align*}
with $u\sim\mathcal{U}(-1,2)$.

\subsection{Transition model}

The model transitions in two steps. The first step is an exchange
between compartments that preserves total population. Denoting with
primes the intermediate state after this first step, we have:
\begin{align*}
s_{t}^{h\prime} & =s_{t-1}^{h}-\oplus e_{t}^{h} & s_{t}^{m\prime} & =s_{t-1}^{m}-\oplus e_{t}^{m}\\
e_{t}^{h\prime} & =e_{t-1}^{h}+\oplus e_{t}^{h}-\oplus i_{t}^{h} & e_{t}^{m\prime} & =e_{t-1}^{m}+\oplus e_{t}^{m}-\oplus i_{t}^{m}\\
i_{t}^{h\prime} & =i_{t-1}^{h}+\oplus i_{t}^{h}-\oplus r_{t}^{h} & i_{t}^{m\prime} & =i_{t-1}^{m}+\oplus i_{t}^{m}-\oplus r_{t}^{m}\\
r_{t}^{h\prime} & =r_{t-1}^{h}+\oplus r_{t}^{h} & r_{t}^{m\prime} & =r_{t-1}^{m}+\oplus r_{t}^{m},
\end{align*}
with the newly exposed, infectious, and recovered populations distributed
as:
\begin{align*}
\oplus e_{t}^{h} & \sim\mathrm{Binomial}(\tau_{t}^{h},\lambda^{h}) & \oplus e_{t}^{m} & \sim\mathrm{Binomial}(\tau_{t}^{m},\lambda^{m})\\
\oplus i_{t}^{h} & \sim\mathrm{Binomial}(e_{t-1}^{h},\delta^{h}) & \oplus i_{t}^{m} & \sim\mathrm{Binomial}(e_{t-1}^{m},\delta^{m})\\
\oplus r_{t}^{h} & \sim\mathrm{Binomial}(i_{t-1}^{h},\gamma^{h}) & \oplus r_{t}^{m} & \sim\mathrm{Binomial}(i_{t-1}^{m},\gamma^{m}),
\end{align*}
for parameters $\lambda^{h}$, $\delta^{h}$, $\gamma^{h}$, $\lambda^{m}$,
$\delta^{m}$, $\gamma^{m}$. The $\tau_{t}^{h}$ gives the number
of susceptible humans bitten by at least one infectious mosquito,
and $\tau_{t}^{m}$ the number of susceptible mosquitos that bite
at least one infectious human:
\begin{align}
\tau_{t}^{h} & \sim\mathrm{Binomial}\left(s_{t-1}^{h},1-\exp(-i_{t-1}^{m}/n_{t-1}^{h})\right)\label{eq:tau_h}\\
\tau_{t}^{m} & \sim\mathrm{Binomial}\left(s_{t-1}^{m},1-\exp(-i_{t-1}^{h}/n_{t-1}^{h})\right).\label{eq:tau_m}
\end{align}
These latter quantities are derived by assuming (a) a $\mathrm{Poisson}(n_{t}^{m})$
number of mosquito blood meals per day with these interactions uniformly
distribution across both humans and mosquitos, (b) that a human is
infected with probability $\lambda^{h}$ if interacting one or more
times with an infectious mosquito, and (c) that a mosquito is infected
with probability $\lambda^{m}$ if interacting one or more times with
an infectious human. Note that the $n_{t-1}^{h}$ appearing in (\ref{eq:tau_h})
is correct, although one may expect to see $n_{t-1}^{m}$ given the
otherwise-symmetry of the equations of this model. In the derivation,
$n_{t-1}^{m}$ also appears in the denominator of both (\ref{eq:tau_h})
and (\ref{eq:tau_m}), but cancels with the Poisson rate parameter
for the number of blood meals, also given by $n_{t}^{m}$ as above.

The second step accounts for births and deaths:
\begin{align*}
s_{t}^{h} & =s_{t}^{h\prime}-\ominus s_{t}^{h}+\oplus n_{t}^{h} & s_{t}^{m} & =s_{t}^{m\prime}-\ominus s_{t}^{m}+\oplus n_{t}^{m}\\
e_{t}^{h} & =e_{t}^{h\prime}-\ominus e_{t}^{h} & e_{t}^{m} & =e_{t}^{m\prime}-\ominus e_{t}^{m}\\
i_{t}^{h} & =i_{t}^{h\prime}-\ominus i_{t}^{h} & i_{t}^{m} & =i_{t}^{m\prime}-\ominus i_{t}^{m}\\
r_{t}^{h} & =r_{t}^{h\prime}-\ominus r_{t}^{h} & r_{t}^{m} & =r_{t}^{m\prime}-\ominus r_{t}^{m},
\end{align*}
with births distributed as
\begin{align*}
\oplus n_{t}^{h} & \sim\mathrm{Binomial}(n_{t}^{h\prime},\nu^{h}) & \oplus n_{t}^{m} & \sim\mathrm{Binomial}(n_{t}^{m\prime},\nu^{m}),
\end{align*}
with parameters $\nu^{h}$ and $\nu^{m}$, and deaths as
\begin{align*}
\ominus s_{t}^{h} & \sim\mathrm{Binomial}(s_{t}^{h\prime},\mu^{h}) & \ominus s_{t}^{m} & \sim\mathrm{Binomial}(s_{t}^{m\prime},\mu^{m})\\
\ominus e_{t}^{h} & \sim\mathrm{Binomial}(e_{t}^{h\prime},\mu^{h}) & \ominus e_{t}^{m} & \sim\mathrm{Binomial}(e_{t}^{m\prime},\mu^{m})\\
\ominus i_{t}^{h} & \sim\mathrm{Binomial}(i_{t}^{h\prime},\mu^{h}) & \ominus i_{t}^{m} & \sim\mathrm{Binomial}(i_{t}^{m\prime},\mu^{m})\\
\ominus r_{t}^{h} & \sim\mathrm{Binomial}(r_{t}^{h\prime},\mu^{h}) & \ominus r_{t}^{m} & \sim\mathrm{Binomial}(r_{t}^{m\prime},\mu^{m}),
\end{align*}
with parameters $\mu^{h}$ and $\mu^{m}$.

\subsection{Observation model}

Observations are of the number of new infectious cases reported at
health centers, aggregated over the time since the last such observation
(this is daily during the peak time of the outbreak and weekly either
side). For times $t\in\left\{ 1,\ldots,T\right\} $ where observations
are available, the observation model is given by
\[
y_{t}\sim\mathrm{Binomial}\left(\sum_{s=t-l_{t}+1}^{t}\oplus i_{s}^{h},\rho\right),
\]
where $l_{t}$ (lag) indicates the number of days since the last observation.
Significant under-reporting of cases is expected, reflected in the
parameter $\rho$.

\subsection{Parameter model}

The following fixed values and priors are assigned to parameters,
translating prior knowledge on rates in \citep{Funk2016a} to prior
knowledge on probabilities here:
\begin{align*}
\nu_{h} & =0 & \nu_{m} & =1/7\\
\mu_{h} & =0 & \mu_{m} & =1/7\\
\lambda_{h} & \sim\mathrm{Beta}(1,1) & \lambda_{m} & \sim\mathrm{Beta}(1,1)\\
\delta_{h} & \sim\mathrm{Beta}\left(\frac{16}{11},\frac{28}{11}\right) & \delta_{m} & \sim\mathrm{Beta}\left(\frac{17}{13},\frac{35}{13}\right)\\
\gamma_{h} & \sim\mathrm{Beta}\left(\frac{13}{9},\frac{23}{9}\right) & \gamma_{m} & =0.
\end{align*}
Birth and death in the human population are assumed to be of minimal
impact over the course of the outbreak, and so their rates are fixed
to zero. The expected lifespan of a mosquito is one week, with birth
and death rates fixed accordingly. Mosquitos do not recover before
death.

Finally, the prior over the reporting probability is
\[
\rho\sim\mathrm{Beta}(1,1).
\]

\subsection{Inference results}

\begin{figure*}
\includegraphics[width=1\textwidth]{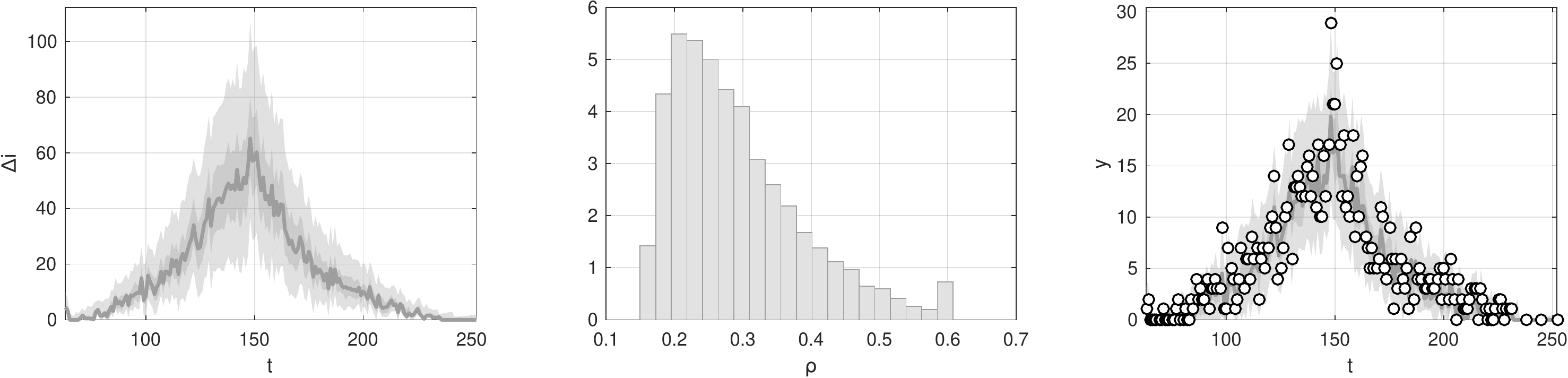}

\caption{Posterior results for the vector-borne disease model example, (left)
posterior distribution of newly infectious cases in humans over time,
$\oplus i_{t}^{h}$, (middle) posterior distribution of the reporting
probability parameter, $\rho$, and (right) posterior predictive distribution
of the number of reported cases, $y$, overlaid with actual observations.
In the left and right plots, the bold line gives the median, darker
shaded region the 50\% credibility interval, and lighter shaded region
the 95\% credibility interval.\label{fig:vbd-posterior}}
\end{figure*}

Inference is performed by drawing 10000 weighted samples, each time
running SMC with 8192 particles. The effective sample size of these
10000 weighted samples is computed to be 2260. Some results are shown
in Figure \ref{fig:vbd-posterior}.

\section{Anglican implementation\label{app:anglican}}

Anglican is a functional probabilistic programming language integrated
with Clojure. Clojure, in turn, is a Lisp dialect which compiles to
Java virtual machine bytecode, enabling reuse of the Java infrastructure.
The Anglican compiler is built with Clojure macros, and compiles Anglican
programs into continuation-passing-style Clojure code. This transformation
enables inference algorithms to affect the control flow and record
information at checkpoints. Manipulations are performed both on the
continuations themselves and on the state, which is passed along as
an argument in each continuation call.

For simplicity, delayed sampling is implemented entirely on top of
the existing Anglican language, leaving the original language constructs
and functionality untouched. A set of new keywords and functions are
added for usage of delayed sampling: \texttt{ds-<name>}, \texttt{ds-value},
and \texttt{ds-observe}. The \texttt{ds-value} and \texttt{ds-observe}
functions loosely correspond to the $\proc{Sample}$ and $\proc{Observe}$
operations in Section \ref{sec:details}, but \texttt{ds-value} also
includes functionality for retrieving values for already-sampled nodes.
The set of \texttt{ds-<name>} functions correspond to the $\proc{Initialize}$
operations in Section \ref{sec:details}, for various probability
distributions, e.g. \texttt{ds-normal}. The delayed sampling graph
is conveniently encoded in the already existing Anglican state.

As an example, consider the following line of code:
\begin{verbatim}
let [x (ds-normal mean sd)]
\end{verbatim}
This binds \texttt{x} to a graph node which is normally distributed
with mean \texttt{mean} and standard deviation \texttt{sd}. To subsequently
introduce another normally distributed graph node with the node \texttt{x}
as mean, one can write
\begin{verbatim}
let [y (ds-normal x sd')]
\end{verbatim}
passing the previous graph node \texttt{x} as a parameter. This will
initialize a conjugate prior relationship between them. If \texttt{y}
is then observed, \texttt{x} will be conditioned on the observed value
of \texttt{y}.

\section{Birch implementation\label{app:birch}}

Birch is a compiled, imperative, object-oriented, generic, and probabilistic
programming language. The latter is its primary research concern.
The Birch compiler uses C++ as a target language.

Delayed sampling has been implemented using the Birch type system.
Special types are used when declaring variables to make them eligible
for delayed sampling. For example, a variable that might ordinarily
be declared to be of type \texttt{Real} may be declared to be of type
\texttt{Random<Real>} to make it eligible for delayed sampling. The
generic class \texttt{Random} implements the behavior required for
delayed sampling, and is specialized into classes that encode distributions
(e.g. \texttt{Gaussian}), then further into classes that encode distributions
with analytical relationships to others (e.g. \texttt{GaussianWithGaussianMean}).
The graph required for delayed sampling is formed implicitly through
objects of these classes and their member attributes.

Birch supports implicit type conversion, compiling directly to the
same feature in C++. These implicit conversions are used to automatically
trigger the $\func{value}$ checkpoint, and are resolved at compile
time. For example, a \texttt{Random<Real>} object may be passed to
a function that requires a \texttt{Real} argument. An implicit conversion
is used to trigger a $\func{value}$ checkpoint, realizing a value
of type \texttt{Real} from the object of type \texttt{Random<Real>}.
In this way, the programmer need not explicitly indicate $\func{value}$
checkpoints.
\end{document}